\documentclass[11pt]{article}
\usepackage{amsthm,amsmath,amssymb,anysize}
\usepackage{graphicx}

\def\qed{\hbox to 0pt{}\hfill$\rlap{$\sqcap$}\sqcup$}

\usepackage[numbers,sort&compress]{natbib}
 \numberwithin{equation}{subsection}

\begin{document}

\title{\textbf{Granule Descriptions based on Compound Concepts\footnote{This paper is supported by the NSF grant of
 Anhui Province (No.1808085MF178), China.} }}
\author{Jianqin Zhou$^{1}$ , Sichun Yang$^{1}$,  Xifeng Wang$^{1}$ and Wanquan Liu$^{2}$ \footnote{Corresponding author.  Email: liuwq63@mail.sysu.edu.cn}\\
\\
\textit{$^{1}$Department of Computer Science,} \\
\textit{Anhui University of   Technology, Ma'anshan 243002,   China}\\
\\
\textit{$^{2}$School of Intelligent Systems Engineering,} \\
\textit{Sun Yat-sen University, Shenzhen 518017,  China}}
% w.liu@curtin.edu.au
\date{ }
\maketitle
\begin{quotation}
\small\noindent

Concise granule descriptions for definable granules and approaching descriptions for indefinable granules are challenging and important issues in granular computing.
The concept with only common attributes has been intensively studied. To investigate the granules with some special needs, we propose a  novel  type of compound concepts in this paper, i.e.,   common-and-necessary concept. Based on the definitions of concept-forming operations, the logical formulas are derived for each of the following types of concepts: formal concept, object-induced three-way concept, object oriented concept and common-and-necessary concept. Furthermore, by utilizing the logical relationship among various concepts, we have derived concise and unified equivalent conditions for definable granules and approaching descriptions for indefinable granules for all four kinds of concepts.

 %. Due to the use of the logical relationship between various concepts, our proof is very easy to understand and the results are particularly concise.

\noindent\textit {Keywords}: Granular computing; Granule description; Approaching description;   Common-and-necessary concept.

\noindent \textit{Mathematics Subject Classification 2010}: 68T30,
68T35

\end{quotation}

 \section{Introduction}

 Granular computing is an emerging computing paradigm, mainly about  representing, reasoning, and processing   granules\cite{Zadeh97}. The granule is defined by Zadeh \cite{ Zadeh97} as a group of objects with some common attributes. The core idea of granular computing is a kind of abstraction. In this sense, granular computing can be used in structured thinking, structured problem solving, and structured information processing. Based on
granular computing, a resilience analysis of critical infrastructures is proposed in \cite{Fujita}. With the notion and framework of three-way granular computing, a multilevel neighborhood sequential decision approach is investigated in \cite{Yang}.

%  The mathematical basis of concept lattice is lattice theory, the visualization tool is the Hasse graph, and the related research methods are abstract algebra, discrete mathematics, data structure and algorithm analysis, fuzzy set \cite{Zadeh}, rough set \cite{Pawlak}, granular computing \cite{Zadeh97}, etc. So far, formal concept analysis has been frequently used in information retrieval \cite{Li}, knowledge discovery \cite{Nguyen}, association analysis \cite{Tu}, recommendation system \cite{Zou} and other fields \cite{ WW}.
 Concept-based granule representation is a popular research topic, and the concept lattice is a key  tool for information processing and analysis. When we study a certain kind of concepts, we first need to consider how to find out all the concepts from the given data. This problem is called concept lattice construction \cite{Andrews17,Andrews,QQW}. Second, in order to better analyze data and save storage space, it is necessary to reduce or simplify the concept lattice \cite{Cao,Ren,Wei}. Furthermore, the nodes of concept lattice can infer from each other, and on this basis, one can extract some inference rules \cite{Li}.

In detail, Zhi and Qi in \cite{Zhi21} observed that the existing types of concepts cannot be used to simultaneously investigate the common attributes and possible attributes of granules.
Thus they proposed the common-possible concepts, where all attributes are from the same collection, and further explored the relationships among the common-possible concepts, formal concepts and object oriented concepts.

In this paper, we will consider a scenario that in an international travel agency, one tour guide can speak several frequently used languages, or can  speak   some infrequently used languages. We first introduce the compound context $(U, A, I, B, J)$, where $A$ is the set of frequently used attributes, and $B$ is the set of infrequently used attributes, on which  \textbf{ common-and-necessary concept} can be defined. Another motivation for the proposed common-and-necessary concept is for a scenario that there usually exist both compulsory courses and elective courses in a school.

%Alternatively, teachers at a school usually pay more attention to two types of students. One kind of students have good test scores, and the other kind of students have poor test scores. Considering this kind of  scenario, we need to define the \textbf{ bipolar concept}.

Li and Liu in \cite{Li20} proposed the concepts of covering element and inserting element for a  granule by which equivalent conditions of definable granules of formal concept and three-way concept were obtained. Such investigation can show the relations clearly for different concepts.

Usually, the definitions of concept operators are given in a description language. However, sometimes the logical formula can better reflect the essence of the problem.  For example,  the essence of formal concept is in the logical formula: $X=a_1 \wedge a_2  \wedge \cdots  \wedge a_k$, and the essence of object oriented concept is in the logical formula: $X=a_1 \vee a_2  \vee \cdots  \vee a_k$, where $(U, A, I)$ is a formal context,  $X\subseteq U$, and $a_1 , a_2  , \cdots a_k\in A$. With above logical formulas, it is easy to decipher that there is a kind of  equivalent relationship between the two concepts. In this paper, the logical formulas are derived for all four kinds of concepts: formal concept, object-induced three-way concept, object oriented concept  and common-and-necessary concept. Thus, by utilizing the logical relationship among
various concepts, and based on the definitions of concept-forming operations, we can derive much concise  and unified equivalent conditions for definable granules and approaching descriptions of indefinable granules for all four kinds of concepts.

In summary, the main contributions in this paper can be summarized as follows.

1. We propose one  new  type  of compound concept, i.e., common-and-necessary concept.

2. The logical formulas have been given for all four kinds of concepts.

3. Utilizing the logical relationship among various concepts, we have derived much concise and unified equivalent conditions for definable granules and approaching descriptions of indefinable granules for all four kinds of concepts.

Finally, the structure of this paper is organized as follows. In Section \ref{s1}, some frequently used definitions and lemmas related
to the main results of this paper are reviewed. In Section \ref{s2}, we present some concise and unified equivalent conditions of definable granules for four kinds of concepts. In Section \ref{s002},  some explicit approaching descriptions for indefinable granules are presented for four kinds of concepts. Finally, in Section \ref{s5},  the paper is concluded with a summary and an outlook for  future work.

\section{Preliminary notions and properties} \label{s1}

%\subsection{Basic notions and properties}\label{sub1.1}

For the convenience of discussion,  we first review some frequently-used notions and their properties related to this paper, such as formal concept, concept-forming operations, logical language, definable granules, and the description of  granule.

 \subsection{ The formal context and its operations  }

 We first present the definition of formal context and its operations as follows.

\textbf{Definition 1}. \cite{Ganter} A triplet  $(U, A, I)$ is called a formal context, if $U =\{x_1, x_2, \ldots $, $x_m\}$, $A=\{a_1, a_2, \ldots , a_n\}$, and $I \subseteq U\times A$ is a binary relation. Here we define each $x_i(i\leq m)$  as an object, and each $a_j(j \leq n)$  as an attribute.  $xIa$ or $(x, a) \in I$ indicates that an object $x\in U$ has the attribute $a\in A$.

In the rest of this paper, $(U, A, I)$ is always used to represent a formal context, and it is a common terminology in granule computing. Some operations for it can be defined as below.

\textbf{Definition 2}. \cite{Ganter} Given a  $(U, A, I)$. For any $X \subseteq U$ and $B \subseteq A$, two concept-forming operations are defined below respectively:

$f : P(U)\rightarrow P(A)$, $f(X) = \{e \in A|\forall x \in X, (x,e) \in I\}$

$g : P(A)\rightarrow P(U)$, $g(B) = \{x \in U | \forall e \in B, (x,e) \in I\}$

With the equipment of above operations, we can give the following definitions of formal concepts and concept lattices.

\textbf{Definition 3}. \cite{Ganter} Given a   $(U, A, I)$. For any $X \subseteq U$ and $B \subseteq A$, if $f(X)=B$ and $g(B)=X$, then we define $(X, B)$ as a formal concept, where $X$ and $B$ are said to be the extent and the intent of $(X, B)$, respectively.

For concepts $(X_1, B_1), (X_2, B_2)$, where $X_1, X_2 \subseteq U$, $ B_1, B_2 \subseteq A$, we can define the partial order as follows:

$(X_1, B_1) \leq (X_2, B_2) \Leftrightarrow X_1 \subseteq X_2\Leftrightarrow B_2 \subseteq B_1$

With the above partial order for forma concept, we can have the following definitions of two set operations:

$(X_1, B_1) \bigwedge (X_2, B_2) = (X_1 \cap X_2, fg(B_1 \cup B_2))$  or $ (X_1 \cap X_2, f(X_1 \cap X_2))  $

$(X_1, B_1) \bigvee (X_2, B_2) = (gf(X_1 \cup X_2), B_1 \cap B_2)$ or $ (g(B_1 \cap B_2), B_1 \cap B_2) $

With all these operations, one can observe that all formal concepts from $(U, A, I)$ would form a complete lattice, which is defined as a concept lattice, which is denoted by $L(U, A, I)$. We have the following properties for the above defined concepts and notations.

\textbf{ Lemma 1}. \cite{Ganter}
 For any $X_1, X_2, X \subseteq U$, $ B_1, B_2, B \subseteq A$, here $(U, A, I)$ is a  context, it is easy to show that the following statements hold:

(1) $X_1\subseteq X_2 \Rightarrow f(X_2) \subseteq f(X_1)$, $\ B_1\subseteq B_2 \Rightarrow g(B_2)\subseteq g(B_1)$;

(2) $X \subseteq gf(X)$, $\ B \subseteq fg(B)$;

(3) $f(X) = fgf(X)$, $\ g(B) = gfg(B)$;

(4) $X \subseteq g(B)\ \Leftrightarrow\ B \subseteq f(X)$;

(5) $f(X_1\cup X_2) = f(X_1)\cap f(X_2), \ g(B_1\cup B_2) = g(B_1)\cap g(B_2)$;

(6) $f(X_1\cap X_2) \supseteq   f(X_1)\cup f(X_2), \ g(B_1\cap B_2) \supseteq g(B_1)\cup g(B_2)$.

\

 In fact, a $(U, A, I)$ can be typically represented by a  table of $0$ and $1$, in which $1$s represents a binary relation between one object (rows) and one attribute (columns). A simple example of a formal context $(U, A, I)$ is illustrated as follows:

 \begin{table*}[ht]
            \begin{center}
                \begin{normalsize}
                    \caption{ $(U, A, I)$}
                    \label{table_tab1}
                    \begin{tabular}
                        {|c|c c  c  c  c|}
                        \hline
                        $U$     & $\ \ \ \ a_1\ \ \ \ \ $       &  $\ \ \ \ a_2\ \ \ \ \ $  &  $\ \ \ \ a_3\ \ \ \ \ $  & $\ \ \ \ a_4\ \ \ \ \ $  &  $\ \ a_5$\ \ \  \\
                       \hline
 1 & 0   & 1  &  1  & 0 &  0  \\

 2 & 1  &  1  & 0 &  0  &  0   \\

 3 & 1   & 0 &  0  &  0  &  0   \\

4 & 0 &  0  &  0  &  0  &   1  \\

5 & 0 &  0  &  0  &  1 &   1  \\

6 & 0 &  0  &  1  &  1 &   1  \\

7 & 1 &  1  &  1  &  0 &   0  \\
\hline
                    \end{tabular}
                \end{normalsize}
            \end{center}
        \end{table*}

%\newpage

With above representation, the formal concepts in  Table \ref{table_tab1}  can be calculated as  given in the following Table \ref{table_2tab}:

 \begin{table*}[ht]
            \begin{center}
                \begin{scriptsize}
                \caption{Formal concepts in  Table 1}
                    \label{table_2tab}
                    \begin{tabular}
                        {c c c c  }
                        \hline
                             &        &  $C_0=(\{1,2,3,4,5,6,7\},\emptyset)$  &       \ \ \  \\

                   &    &     &        \\

 $C_4=(\{2,3,7\},\{a_1\}) $ & $C_3=(\{1,2,7\},\{a_2\}) $   & $C_2=(\{1,6,7\},\{a_3\}) $  &  $C_1=(\{4,5,6\},\{a_5\}) $     \\

   &    &     &         \\

  & $C_5=(\{2,7\},\{a_1, a_2\}) $  &  $C_6=(\{1,7\},\{a_2, a_3\}) $   & $C_7=(\{5,6\},\{a_4, a_5\}) $      \\

  &    &     &         \\

  &  $C_9=(\{7\},\{a_1, a_2, a_3\}) $ &    &  $C_8=(\{6\},\{a_3, a_4, a_5\}) $    \\

   &    &     &         \\

 &  &    $C_{10}=(\emptyset,\{a_1, a_2, a_3 ,a_4, a_5\}) $   &      \\
\hline

                    \end{tabular}
                \end{scriptsize}
            \end{center}
        \end{table*}

  One can visualize a formal concept in a  table of $0$ and $1$  as a closed rectangle of $1$s, where the rows and columns  are not necessarily contiguous \cite{Andrews}.
   We define the cell of the $i$th row and $j$th column as $(i,j)$. Thus in Table \ref{table_tab1}, $(5,4)$, $(5,5)$, $(6,4)$ and $(6,5)$ would form the concept $C_7$, and $C_7$ is a rectangle with height $2$ and width $2$. Similarly $(6,3)$, $(6,4)$ and $(6,5)$ form the concept $C_8$, and $C_8$ is a rectangle of height $1$ and width $3$.  $(1,3)$, $(6,3)$ and $(7,3)$ form the concept $C_2$, and $C_2$ is a rectangle of height $3$ and width $1$. Here $(1,3)$ and $(6,3)$ are not contiguous. Next, we first investigate the definable granule.

\subsection{Definable granule and its description}

Given a formal context $(U, A, I)$. We  present a logical language used for describing a granule $X\subseteq U$. For any $b\in B \subseteq A$, we can define $b$ as an atomic formula. By joining all the atomic formulas in $B$ together with the connective $\wedge$, we can obtain a compound formula  $\wedge B =\bigwedge_{b\in B}b $.

Furthermore, if the object $x\in U$ has an attribute $b$, then we say that $x$  satisfies the atomic formula $b$, denoted by $x\mapsto b$. Obviously, if $x \mapsto b$ for any $b \in B$, then $x$ is said to satisfy the compound formula $\wedge B$, denoted by $x \mapsto \wedge B$. Thus the semantics of $\wedge B$ is defined as the set of the
objects satisfying $\wedge B$ as defined below,

\[  m(\wedge B) = \{ x\in U |\forall b\in B, x\mapsto b \} \]

  Given a  formal context $(U, A, I)$ and $B\subseteq A$ with $B\neq \emptyset$. Then we have $m(\wedge B)= g(B)$   \cite{Zhi16}.

With the semantics operator $m$ defined above, we now give the  definition  of a $\wedge$-definable
granule.

\textbf{Definition 4}. \cite{Zhi16}  Given a   $(U, A, I)$ and $X\subseteq U$. If there exists  $B\subseteq A, B\neq \emptyset$, such that $m(\wedge B)= X$, then we say that $X$ is $\wedge$-definable and $\wedge B$ is
 a description of $X$, denoted by $d(X)=\wedge B$.

The following proposition is an immediate result derived from the above definition.

\textbf{ Proposition 1}. \cite{Zhi16}  Given a formal context  $(U, A, I)$, $Y\subseteq U$ and  $f(Y)\neq \emptyset$.
 If $Y$ is the extent of a  formal concept, then the granule $Y$ must be
$\wedge$-definable and $d(Y)=\wedge f(Y)$.

\

For the example in Table \ref{table_tab1}, we can rewrite \[a_1=\begin{pmatrix} 0\\ 1\\ 1\\ 0\\ 0\\ 0\\ 1 \end{pmatrix}, \ a_2=\begin{pmatrix} 1\\ 1\\ 0\\  0\\ 0\\ 0\\ 1\end{pmatrix}\] as column vectors, then $a_1 \wedge a_2 =\begin{pmatrix} 0\\ 1\\ 0\\ 0\\ 0\\ 0\\ 1\end{pmatrix}$,
 where $a_1, a_2\in A$. At the same time, $x_2=( 1,1,0,0,0)$, $x_7=( 1,1,1, 0,0)$, then $x_2 \wedge x_7 =(1,1,0,0,0)$, where $x_2, x_7\in U$.
 Thus one has $X=\{x_2, x_7\}=a_1 \wedge a_2$, and $X$ is $\wedge$-definable, $d(X)= a_1 \wedge a_2$, $\{x_2, x_7\}=m(a_1 \wedge a_2)$. Also, $C_5=(\{x_2,x_7\},\{a_1, a_2\}) $ is a formal concept.

 In essence, $X\subseteq U$ is
$\wedge$-definable $\Longleftrightarrow$ there exist $a_1 , a_2  , \cdots a_k\in A$, such that $X=a_1 \wedge a_2  \wedge \cdots  \wedge a_k$.
 It is worth noticing that $\{x_6\}= a_3 \wedge a_4$, also $\{x_6\}= a_3 \wedge a_4 \wedge a_5$. So $d(\{x_6\})= a_3 \wedge a_4$ and $d(\{x_6\})= a_3 \wedge a_4 \wedge a_5$. Based on the
 above observation, one can see that the description of  $X\subseteq U$ may be not unique.
 %Thus we need to provide the equivalent descriptions for definable granules.

\section{Equivalent conditions of definable granules} \label{s2}

 With the concept operations, we can present some unified equivalent conditions of definable granules for each of the following types of concepts: formal concept, three-way object-induced concept, object oriented concept  and common-and-necessary concept.

\subsection{$\wedge$-definable granules}\label{sub3.1}

In \cite{Li20}, the notion of covering element was proposed, and the following proposition was obtained.

\textbf{Proposition 2}. \cite{Li20} Given a formal context $(U, A, I)$, $X\subseteq U$ and  $f(X)\neq \emptyset$.  Then
$X$ is  $\wedge$-definable $\Longleftrightarrow$ $X$ does not have any covering element $y\notin X$, such that $f(X)\subseteq f(y)$.

Based on concept-forming operations of Definition 2, we give the following theorem.

\textbf{Theorem 1}.  Given a  formal context $(U, A, I)$, $X\subset U$ and  $f(X)\neq \emptyset$.  Then
$X$ is   $\wedge$-definable $\Longleftrightarrow$ $X=gf(X)$.

\begin{proof}
$\Longrightarrow$ By Definition 4, there exists  $B\subseteq A, B\neq \emptyset$, such that $m(\wedge B)= X   \Longleftrightarrow g(B)=X$.
By Lemma 1.(3), $X=g(B)=gfg(B)=gf(X)$.

$\Longleftarrow$ Let $B=f(X)$. Then $g(B)=gf(X)=X$. So $m(\wedge B)= X$, thus $X$ is   $\wedge$-definable.
\end{proof}

By Lemma 1.(2),  $X \subseteq gf(X)$.  If $X \subset  gf(X)$, then there exists $y\in gf(X)$ and $y\notin X$, such that $f(y) \supseteq fgf(X)= f(X) $.
Thus $y$ must be a covering element. Therefore, the conditions of Theorem 1 and Proposition 2 are equivalent.

If $X$ is   $\wedge$-definable, $B=f(X)=\{ a_1 , a_2  , \cdots a_k  \}$ $\Longrightarrow$ $X=g(B) \Longrightarrow X=a_1 \wedge a_2  \wedge \cdots  \wedge a_k$. If $Y$ is   $\wedge$-definable, $C=f(Y)=\{ b_1 , b_2  , \cdots b_j  \}$ $\Longrightarrow$ $Y=g(C) \Longrightarrow Y=b_1 \wedge b_2  \wedge \cdots  \wedge b_j$.  Thus
\[ X\bigcap Y=  a_1 \wedge a_2  \wedge \cdots  \wedge a_k \wedge b_1 \wedge b_2  \wedge \cdots  \wedge b_j\]
\[g(f(X)\bigcup f(Y))=g(\{  a_1 , a_2  , \cdots a_k,  b_1 , b_2  , \cdots b_j  \})= X\bigcap Y \]
 So the following Proposition 3 is obvious.

\textbf{Proposition 3}. \cite{Li20} Given a formal context  $(U, A, I)$, $X,Y\subset U$ and  $f(X)\neq \emptyset$, $f(Y)\neq \emptyset$.  If $X$ and $Y$ are  $\wedge$-definable, then the granule $X\bigcap Y$ must be   $\wedge$-definable with $d(X\bigcap Y)= \wedge ( f(X)\bigcup f(Y)) $.

However, as $ X\bigcup Y=  a_1 \wedge a_2  \wedge \cdots  \wedge a_k \vee b_1 \wedge b_2  \wedge \cdots  \wedge b_j $,  $ X\bigcup Y$ maybe not $\wedge$-definable.

\subsection{($\wedge, \wedge, \neg$)-definable granules}\label{sub3.02}

Based on Definition 3 in \cite{Li20}, from a  formal context $(U, A, I)$, we can build a compound context $(U, A, I, B, J)$, where $a\in A$ and  $(x,a)\in I \Longleftrightarrow  b\in B$ and $ (x,b)\notin J$. In this compound context, $A,B$ and $I,J$ are associated with each other.

In the rest of this paper, $(U, A, I,  B, J)$ is always used to represent a compound context.

\

For any $X \subseteq U$ and $C \subseteq A\bigcup B$,
we further extend the concept-forming operations $f$ and $g$ as follows:

\[f : P(U)\rightarrow P(A\bigcup B), f(X) = \{e \in A \bigcup B |\forall x \in X, (x,e) \in I \bigcup J\}\]

\[g : P(A\bigcup B)\rightarrow P(U),  g(C) = \{x \in U | \forall e \in  C, (x,e) \in (I\bigcup J)  \}   \]

Evidently, $f$ is the combination of $f^+$ and $f^-$ in \cite{Li20}, and $g$ is the combination of $g^+$ and $g^-$ in \cite{Li20}.
%It is easy to show that Lemma 1 still holds for $f$, $g$ in the compound context $(U, A, I, B, J)$.

  Given a compound context $(U, A, I, B, J)$. For any $X \subseteq U$ and $C \subseteq A\bigcup B$, if $f(X)=C$ and $g(C)=X$, then we say that  $(X, C)$   is   an object-induced three-way concept in \cite{Qi14}. As  involving both the logic connectives $\wedge$ and $\neg$, we define it as a compound concept.

Based on Definition 4 in \cite{Li20}, we give a new notion of ($\wedge, \wedge, \neg$)-definable.

\textbf{Definition 5}. Given a compound context  $(U, A, I, B, J)$ and $X\subseteq U$, if there exist $C\neq \emptyset$,   $C \subseteq A\cup B$, such that
$g(C) = X$, then we say that the granule $X$ is  ($\wedge, \wedge, \neg$)-definable, and $d(X)= \wedge(C)$.

Similar to $\wedge$-definable, we can obtain the following concise result about ($\wedge, \wedge, \neg$)-definable.

\textbf{Theorem 2}.  Given a compound context $(U, A, I, B, J)$, $X\subset U$ and  $f(X)\neq \emptyset$.  Then
$X$ is   ($\wedge, \wedge, \neg$)-definable $\Longleftrightarrow$ $X=gf(X)$.

%\begin{proof}
% Necessity. By Definition 5, there exist  $C\neq \emptyset$, $D\neq \emptyset$, $C \subseteq A$ and $D \subseteq B$, such that
%$g_T(C\bigcup D) = X$ $\Longrightarrow X= g(C)\bigcap g(D)$, where $g$ is defined in Definition 2.
%
% By  Lemma 1.(5),
%$  X= g(C)\bigcap g(D)=g(C\bigcup D)$.
%
%By Lemma 1.(3), $X=g(C\bigcup D)=gfg(C\bigcup D)=gf(X) =g_Tf(X)$.
%
%
%Sufficiency. Let $C\bigcup D=f(X)$, where  $f(X)\bigcap A=C$,  $f(X)\bigcap B=D$. Then $g_T(C\bigcup D)=g_Tf(X)=X$. Thus $X$ is   ($\wedge,\wedge,\neg$)-definable.
%\end{proof}

One can observe that Theorem 2 in \cite{Li20} also gave a equivalent condition of ($\wedge,\wedge,\neg$)-definable
granule only for the  case of $X\subseteq U$, $X$ is $\wedge$-indefinable in $(U, A, I)$.  In comparison with the above result, Theorem 2 in this paper is significant extension in comparison to Theorem 2 in \cite{Li20}

\

Furthermore, if $X$ is ($\wedge,\wedge,\neg$)-definable, $f(X)=\{ a_1 , a_2  , \cdots a_k  \}$ $\Longrightarrow$ $X=gf(X)$ $\Longrightarrow$
$$X=a_1 \wedge a_2  \wedge \cdots  \wedge a_k$$
where $a_1 , a_2  , \cdots a_k\in A\bigcup B$.%,  $\{ a_1 , a_2  , \cdots a_k  \}\bigcap A \neq \emptyset$ and   $\{ a_1 , a_2  , \cdots a_k  \}\bigcap B \neq \emptyset$.

Also, if $Y$ is ($\wedge,\wedge,\neg$)-definable, $f(Y)=\{ b_1 , b_2  , \cdots b_j  \}$ $\Longrightarrow$ $Y=gf(Y)$ $\Longrightarrow$
$$ Y=b_1 \wedge b_2  \wedge \cdots  \wedge b_j$$
where $b_1 , b_2  , \cdots b_j\in A\bigcup B$.%, $\{ b_1 , b_2  , \cdots b_j  \}\bigcap A \neq \emptyset$ and   $\{ b_1 , b_2  , \cdots b_j  \}\bigcap B \neq \emptyset$.

Therefore,
\[ X\bigcap Y=  a_1 \wedge a_2  \wedge \cdots  \wedge a_k \wedge b_1 \wedge b_2  \wedge \cdots  \wedge b_j\]
\[g(f(X)\bigcup f(Y))=g(\{  a_1 , a_2  , \cdots a_k,  b_1 , b_2  , \cdots b_j  \})= X\bigcap Y \]

By combining the above results, one can see the following Proposition 4 is obvious.

\textbf{Proposition 4}.   Given a compound context $(U, A, I, B,J)$, $X,Y\subset U$ and  $f(X)\neq \emptyset$, $f(Y)\neq \emptyset$.  If $X$ and $Y$ are ($\wedge,\wedge,\neg$)-definable, then the granule $X\bigcap Y$ must be ($\wedge,\wedge,\neg$)-definable with $d(X\bigcap Y)= \wedge ( f(X)\bigcup f(Y)) $.

However, as $ X\bigcup Y=  a_1 \wedge a_2  \wedge \cdots  \wedge a_k \vee b_1 \wedge b_2  \wedge \cdots  \wedge b_j $,  $ X\bigcup Y$ may be not ($\wedge,\wedge,\neg$)-definable.

As Proposition 11 in \cite{Li20} only discuss the  case of $X\subseteq U$, $X$ is $\wedge$-indefinable in $(U, A, I)$,
which indicates that Proposition 4 here is broader than Proposition 11 in \cite{Li20}.

%As $ X\bigcup Y=  a_1 \wedge a_2  \wedge \cdots  \wedge a_k \vee b_1 \wedge b_2  \wedge \cdots  \wedge b_j $,  $ X\bigcup Y$ maybe not ($\wedge,\wedge,\neg$)-definable.

\

For example, from a formal context $(U, A, I)$ in Table \ref{table_tab1}, we can create a compound context  $(U, A, I, B, J)$ described as below in Table \ref{table_tab0033},  where $a\in A \Leftrightarrow b\in B$, and $(x,a)\in I \Leftrightarrow (x,b)\notin J$.

 \begin{table*}[ht]
            \begin{center}
                \begin{normalsize}
                    \caption{ $(U, A,I, B, J)$}
                    \label{table_tab0033}
                    \begin{tabular}
                        {|c|c c  c  c  c   c c  c  c  c |}
                        \hline
                        $U$    & $\ \  \ a_1\     \ \ $      &  $\    \ a_2\ \    \ $  &  $\  \ a_3\    \ \ $  & $\   \ a_4\     \ $  &  $\ \ a_5$\   \ & $\ \  \ b_1\    \ \ $       &  $\   \ b_2\    \ \ $  &  $\  \ b_3\   \ \ $  & $\    \ b_4\   \ $  &  $\ \ b_5$\   \  \\
                       \hline
 1 & 0   & 1  &  1  & 0 &  0  & 1   & 0  &  0  & 1 &  1  \\

 2 & 1  &  1  & 0 &  0  &  0  & 0  &  0  & 1 &  1  &  1   \\

 3 & 1   & 0 &  0  &  0  &  0 & 0   & 1 &  1  &  1  &  1   \\

4 & 0 &  0  &  0  &  0  &   1 & 1 &  1  &  1  &  1  &   0  \\

5 & 0 &  0  &  0  &  1 &   1  & 1 &  1  &  1  &  0 &    0  \\

6 & 0 &  0  &  1  &  1 &   1  & 1 &  1  &  0  &  0 &    0  \\

7 & 1 &  1  &  1  &  0 &   0 & 0 &  0  &  0  &  1 &   1  \\
\hline
                    \end{tabular}
                \end{normalsize}
            \end{center}
        \end{table*}

        From Table \ref{table_2tab}, $C_5=(\{2,7\},\{a_1, a_2\}) $  is a concept over $(U, A, I)$. However, $(\{2,7\}$, $\{a_1, a_2, b_4,b_5\}) $
        is a concept over $(U, A, I, B,J)$ and $ \{2,7\} = a_1\wedge a_2\wedge b_4\wedge b_5$, which is not covered by Theorem 2 in \cite{Li20}.

Note that $ \{2,3\} = a_1\wedge b_3\wedge b_4 \wedge b_5$, by Proposition 4, $\{2,3\}\bigcap \{2,7\} = \{2\}= a_1\wedge a_2\wedge b_3\wedge b_4\wedge b_5$,
which is still ($\wedge,\wedge,\neg$)-definable. Next, we will describe  the important   $\vee$-definable granules.

\subsection{$\vee$-definable granules}

 To investigate the other aspects of granules,  we next present the possibility
operator $(\cdot)^\diamond$ and necessity operator $(\cdot)^\Box$.
%For convenience, $(\cdot)^\Box$ is denoted  as $f_\vee$, and  $(\cdot)^\diamond$ is denoted  as $g_\vee$ in the remaining part of this paper.

\textbf{Definition 6}. \cite{ Zhi18} Given a formal context  $(U, A, I)$,   $X \subseteq U$ and $B \subseteq A$. Then two concept-forming operations are defined below respectively:

\[f_\vee : P(U)\rightarrow P(A), f_\vee(X) = X^\Box;\ g_\vee : P(A)\rightarrow P(U),  g_\vee(B) = B^\diamond\]
where
\[X^\Box=\{e \in A| Ie\subseteq X \}=\{e \in A|\forall x \in U, (x,e) \in I  \Rightarrow x \in X \}\]
\[B^\diamond =\bigcup\limits_{ e \in B}Ie=\{x \in U |xI\cap B\ne \emptyset\}=\{x \in U | \exists e \in B, (x,e) \in I\}\]
\[   xI=\{  e\in A | (x,e)\in I     \}, Ie=\{  x\in U | (x,e)\in I     \}  \]

Based on the above two concept-forming operations, one can present the object oriented concept.
For any $X \subseteq U$ and $B \subseteq A$, if $f_\vee(X)=B$ and $g_\vee(B)=X$, then  $(X, B)$   is called an object oriented  concept.

\textbf{Definition 7}. \cite{Zhi18}  Given a formal context $(U, A, I)$ and $X\subseteq U$. If there exists  $B\subseteq A, B\neq \emptyset$, such that $g_\vee( B)= X$, then we say that  $X$ is  $\vee$-definable and  $d(X)=\vee B$.

 In essence, $X\subseteq U$ is
$\vee$-definable $\Longleftrightarrow$ there exist $a_1 , a_2  , \cdots a_k\in A$, such that
$$X=a_1 \vee a_2  \vee \cdots  \vee a_k$$

Furthermore, from a formal context $(U, A, I)$, we can create a  context $(U,  B, J)$, where $a\in A \Leftrightarrow b\in B$, and $(x,a)\in I \Leftrightarrow (x,b)\notin J$. In this case, we have
 \[ X=a_1 \vee a_2  \vee \cdots  \vee a_k \Longleftrightarrow U\backslash X= b_1 \wedge b_2  \wedge \cdots  \wedge b_k\]

Therefore, $X$ is   $\vee$-definable over  $(U,  A, I)$  $\Longleftrightarrow$  $U\backslash X$ is  $\wedge$-definable over  $(U,  B, J)$.
The following Theorem 3 is immediate from Theorem 1.

\textbf{Theorem 3}.  Given a formal context  $(U, A, I)$ and $X\subset U$.  Then
$X$ is   $\vee$-definable $\Longleftrightarrow$ $U\backslash X=gf(U\backslash X)$ over formal context $(U,  B, J)$, where $a\in A \Leftrightarrow b\in B$, and $(x,a)\in I \Leftrightarrow (x,b)\notin J$.

\

If $X$ is $\vee$-definable, one will have $f(U\backslash X)=\{ a_1 , a_2  , \cdots a_k  \}$ $\Longrightarrow$ $U\backslash X=gf(U\backslash X) \Longrightarrow U\backslash X=a_1 \wedge a_2  \wedge \cdots  \wedge a_k$.

Similarly, if $Y$ is $\vee$-definable, $f(U\backslash Y)=\{ b_1 , b_2  , \cdots b_j  \}$ $\Longrightarrow$ $U\backslash Y=gf(U\backslash Y) \Longrightarrow U\backslash Y=b_1 \wedge b_2  \wedge \cdots  \wedge b_j$.  By combining the above two results, one will have

\[(U\backslash X)\bigcap (U\backslash Y)=  a_1 \wedge a_2  \wedge \cdots  \wedge a_k \wedge b_1 \wedge b_2  \wedge \cdots  \wedge b_j\]
\[g(f(U\backslash X)\bigcup f(U\backslash Y))=g(\{  a_1 , a_2  , \cdots a_k,  b_1 , b_2  , \cdots b_j  \})=(U\backslash X)\bigcap (U\backslash Y) \]
 So the following Proposition 5 is obvious.

\textbf{Proposition 5}.   Given a formal context  $(U, A, I)$ and $X,Y\subset U$.  If $X$ and $Y$ are   $\vee$-definable, then
$U\backslash (U\backslash X)\bigcap (U\backslash Y)$ is   $\wedge$-definable, and
 $d((U\backslash X)\bigcap(U\backslash Y))$ \ \ \ \
 $=$  $\wedge ( f(U\backslash X)\bigcup f(U\backslash Y)) $.
\\

As $(U\backslash X)\bigcup(U\backslash Y)=  a_1 \wedge a_2  \wedge \cdots  \wedge a_k \vee b_1 \wedge b_2  \wedge \cdots  \wedge b_j $,  $ (U\backslash X)\bigcup (U\backslash Y)$ may be not the $\wedge$-definable.

\

In fact, similar to Theorem 1, it is easy to obtain the following Proposition 6, which is equivalent to Theorem 3.

\textbf{Proposition 6}.  Given a formal context  $(U, A, I)$, $X\subset U$ and  $f_{\vee}(X)\neq \emptyset$.  Then
$X$ is   $\vee$-definable $\Longleftrightarrow$ $X=g_{\vee}f_{\vee}(X)$.

\

For example, from a formal context  $(U, A, I)$ in Table \ref{table_tab1}, we can create a formal context $(U,  B, J)$, given below in Table \ref{table_tab03},  where $a\in A \Leftrightarrow b\in B$, and $(x,a)\in I \Leftrightarrow (x,b)\notin J$.

 \begin{table*}[ht]
            \begin{center}
                \begin{normalsize}
                    \caption{Formal context $(U, B, J)$}
                    \label{table_tab03}
                    \begin{tabular}
                        {|c|c c  c  c  c|}
                        \hline
                        $U$     & $\ \ \ \ b_1\ \ \ \ \ $       &  $\ \ \ \ b_2\ \ \ \ \ $  &  $\ \ \ \ b_3\ \ \ \ \ $  & $\ \ \ \ b_4\ \ \ \ \ $  &  $\ \ b_5$\ \ \  \\
                       \hline
 1 & 1   & 0  &  0  & 1 &  1  \\

 2 & 0  &  0  & 1 &  1  &  1   \\

 3 & 0   & 1 &  1  &  1  &  1   \\

4 & 1 &  1  &  1  &  1  &   0  \\

5 & 1 &  1  &  1  &  0 &    0  \\

6 & 1 &  1  &  0  &  0 &    0  \\

7 & 0 &  0  &  0  &  1 &   1  \\
\hline
                    \end{tabular}
                \end{normalsize}
            \end{center}
        \end{table*}

On one hand,
$(\{1,4,5,6,7\},\{a_3, a_4, a_5\}) $  is an object oriented concept over $(U, A, I)$ in Table \ref{table_tab1}, and $\{1,4,5,6,7\} = a_3\vee a_4\vee a_5$.
On the other hand, $(\{2,3\},\{b_3, b_4, b_5\}) $  is a formal concept over formal context $(U, B, J)$ in Table \ref{table_tab03}, and
 $ U\backslash \{1,4,5,6,7\}=   \{2,3\} = b_3\wedge b_4\wedge b_5$.

Similarly,
$(\{1,2,5,6,7\},\{a_2, a_3, a_4\}) $  is an object oriented concept over   $(U, A, I)$ in Table \ref{table_tab1}, and $\{1,2,5,6,7\} = a_2 \vee a_3\vee a_4 $.
At the same time, $(\{3,4\},\{b_2, b_3, b_4\}) $  is a formal concept over formal context $(U, B, J)$ in Table \ref{table_tab03}, and
 $ U\backslash \{1,2,5,6,7\}=   \{3,4\} = b_2\wedge b_3\wedge b_4$.

 As $\{2,3\}\cap \{3,4\}= \{3\} = b_2\wedge b_3\wedge b_4 \wedge b_5$, from Proposition 5, $\{1,2, 4,5,6,7\} = a_2 \vee a_3\vee a_4\vee a_5$.

 As there are many fast algorithms for computing formal concepts \cite{Andrews, Andrews17},  so these algorithms can be used  for computing the object oriented concepts and    attribute oriented concepts \cite{Yao}.

\

\subsection{($\wedge, \wedge, \vee$)-definable granules}

In \cite{Zhi21}, the common-possible concept was proposed for concurrently investigating the common attributes and possible attributes of granules, where all attributes are from the   same collection. In this subsection, we still consider the scenario that in  an international travel agency, one guide can speak several  frequently used languages, or can  speak   infrequently used languages.

 We   introduce a  compound context $(U, A, I, B, J)$, where $A$ is the set of frequently used attributes, and $B$ is the set of infrequently used attributes. For any $X \subseteq U$ and $E \subseteq A\bigcup B$, we further introduce the concept-forming operations $f_{CN}$ and $g_{CN}$ as follows:

\[f_{CN} : P(U)\rightarrow P(A\bigcup B),\]{%\footnotesize
\[ f_{CN}(X) = \max\limits_{  a_1 \wedge a_2  \wedge \cdots  \wedge a_k\supseteq X} \{    a_1 , a_2  , \cdots  , a_k  \in A  \} \bigcup \min\limits_{  b_1 \vee b_2  \vee \cdots  \vee b_j \supseteq X} \{b_1 , b_2  , \cdots  , b_j \in B\}\]}
\[g_{CN} : P(A\bigcup B)\rightarrow P(U),\]
\[ g_{CN}(E) = \{u \in U | \forall e \in A\bigcap E, (u,e) \in I  \}   \bigcap   ( B\bigcap E)^{\diamond}  \]
where $(B\bigcap E)^{\diamond}=\{u \in U | \exists e \in B\bigcap E, (u,e) \in   J\}$.
%where $f_2(X_2)=\{e \in B|\forall g \in U, (g,e) \in J  \Rightarrow g \in X_2 \}$ and $X_2=\{g \in U | \exists e \in f_2(X_2), (g,e) \in   J \}$

 Given a compound context $(U, A, I, B, J)$, for any $X \subseteq U$ and $E \subseteq A\bigcup B$, if $f_{CN}(X)=E$ and $g_{CN}(E)=X$, then the pair $(X, E)$   is called a {\bf common-and-necessary  concept}. As  this concept also includes  two logic connectives $\wedge$ and $\vee$, we define it as a compound concept.

Furthermore, we give the definition of ($\wedge, \wedge, \vee$)-definable.

\textbf{Definition 8}. Given a compound context $(U, A, I, B, J)$ and $X\subseteq U$,
if there exist $ C\neq \emptyset$, $ D\neq \emptyset$, $C \subseteq A$ and $D \subseteq B$, such that
$g_{CN}(C\bigcup D) = X$, then we say that the granule $X$ is  ($\wedge,\wedge, \vee$)-definable, and $d(X)= \wedge(C) \wedge\vee(D)$.

Based on the definitions of $f_{CN}$ and $g_{CN}$,  we prove the following  theorem.

\textbf{Theorem 4}.  Given a compound context $(U, A, I, B, J)$, $X\subset U$ and  $f_{CN}(X)\bigcap A\neq \emptyset$, $f_{CN}(X)\bigcap B\neq \emptyset$.  Then
$X$ is   ($\wedge, \wedge, \vee  $)-definable $\Longleftrightarrow$ $X=g_{CN}f_{CN}(X)$.

\begin{proof}
Necessity. By Definition 8, there exist $ C\neq \emptyset$, $ D\neq \emptyset$, $C \subseteq A$ and $D \subseteq B$, such that
$g_{CN}(C\bigcup D) = X$.
By the definition of $g$ in Definition 2, let $X_1=g(C)$, $X_2=D^{\diamond}$. Then $X_1\bigcap X_2= X$.

By the definition of $f_{CN}$, and note that $A\bigcap B=\emptyset$, $ f_{CN}(X)\bigcap A$ is the maximum over $A$ and $f_{CN}(X)\bigcap B$ is the minimum over $B$. Then,

  $X_1\bigcap X_2= X \Longrightarrow C\subseteq (f_{CN}(X)\bigcap A)$  and $D\supseteq (f_{CN}(X)\bigcap B)$

$$\Longrightarrow g_{CN}((f_{CN}(X)\bigcap A ) \bigcup (f_{CN}(X)\bigcap B ))\subseteq g(C)\bigcap D^{\diamond} = X_1\bigcap X_2 =X$$

On the other hand, by the definition of $f_{CN}$,
$$  g(f_{CN}(X)\bigcap A)\supseteq X$$ and
$$ (f_{CN}(X)\bigcap B)^{\diamond}\supseteq X$$

$$ \Longrightarrow g_{CN}((f_{CN}(X)\bigcap A ) \bigcup (f_{CN}(X)\bigcap B ))\supseteq X$$

  Therefore, $g_{CN}f_{CN}(X)=X$.

Sufficiency. Let $C\bigcup D=f_{CN}(X)$, where  $f_{CN}(X)\bigcap A=C$,  $f_{CN}(X)\bigcap B=D$. Then $g_{CN}(C\bigcup D)=g_{CN}f_{CN}(X)=X$. Thus $X$ is   ($\wedge,\wedge, \vee$)-definable.
\end{proof}

In fact, $X$ is   ($\wedge,\wedge, \vee$)-definable $\Longleftrightarrow X= a_1 \wedge a_2  \wedge \cdots  \wedge a_k \wedge( b_1 \vee b_2  \vee \cdots  \vee b_j) $, where $a_1 , a_2  , \cdots  , a_k\in A$ and $b_1 , b_2  , \cdots  , b_j \in B$.

 \begin{table*}[ht]
            \begin{center}
                \begin{normalsize}
                    \caption{   $(U, A,I, B, J)$}
                    \label{table_tab005}
                    \begin{tabular}
                        {|c|c c  c  c  c   c c  c  c   |}
                        \hline
                        $U$    & $\ \  \ a_1\     \ \ $      &  $\    \ a_2\ \    \ $  &  $\  \ a_3\    \ \ $  & $\   \ a_4\     \ $  &  $\ \ a_5$\   \ & $\ \  \ b_1\    \ \ $       &  $\   \ b_2\    \ \ $  &  $\  \ b_3\   \ \ $  & $\    \ b_4\   \ $    \\
                       \hline
 1 & 0   & 1  &  1  & 0 &  0  & 1   & 0  &  0  & 0    \\

 2 & 1  &  1  & 0 &  0  &  0  & 0  &  0  & 1 &  1      \\

 3 & 1   & 0 &  0  &  0  &  0 & 0   & 0 &  1  &  1      \\

4 & 0 &  0  &  0  &  0  &   1 & 0 &  0  &  1  &  0     \\

5 & 0 &  0  &  0  &  1 &   1  & 0 &  0  &  1  &  0    \\

6 & 0 &  0  &  1  &  1 &   1  & 1 &  0  &  0  &  0    \\

7 & 1 &  1  &  1  &  0 &   0 & 1 &  1  &  0  &  1    \\
\hline
                    \end{tabular}
                \end{normalsize}
            \end{center}
        \end{table*}

For example, in Table \ref{table_tab005},   $ \{2,3,7\} =   a_1    $, $ \{2,3,7\} =   b_2\vee b_4 $, thus $ \{2,3, 7\} =a_1\wedge (  b_2\vee b_4)  $, where $f_{CN}(\{2,3,7\})=\{a_1,  b_2, b_4  \}$, and  $g_{CN}f_{CN}(\{2,3,7\})= g_{CN}(\{ a_1,  b_2, b_4   \})= \{2,3, 7\}$. Thus $X=\{2,3, 7\}$ is   ($\wedge,\wedge, \vee$)-definable.
However,  we also have $ \{2,3, 7\} =a_1\wedge   b_4  $, thus   $d(\{2,3, 7\})$   is not  unique.

Similarly,  $g_{CN}f_{CN}(\{1,6,7\})= g_{CN}(\{ a_3,  b_1, b_2   \})= \{1,6, 7\}$. Thus $X=\{1,6, 7\}$ is   ($\wedge,\wedge, \vee$)-definable.

Also,  $g_{CN}f_{CN}(\{2,3\})= g_{CN}(\{ a_1,  b_3   \})= \{2,3\}$. Thus $X=\{2,3\}$ is   ($\wedge,\wedge, \vee$)-definable. Note that here $X\subset a_1=\{2,3,7\}$ and $X\subset b_3=\{2,3,4,5\}$.

\

 \begin{table*}[ht]
            \begin{center}
                \begin{normalsize}
                    \caption{Student scores}
                    \label{table_course}
                    \begin{tabular}
                        {|c|c c  c  c  c  c|}
                        \hline
                        $Name$     & $\ \ \ \ c_1\ \ \ \ \ $       &  $\ \ \ \ c_2\ \ \ \ \ $  &  $\ \ \ \ c_3\ \ \ \ \ $  & $\ \ \ \ c_4\ \ \ \ \ $  &  $\ \ ec_1$\ \ \ &  $\ \ ec_2$\ \ \ \\
                       \hline
 Peter & 1   & 1  &  1  & 1  &  0   &  0 \\

 John & 1 &  0  &  1   &  1    &  1    &  0 \\

 Grace & 1   & 1 &  1  &  1  &  0  &  1 \\

Jenny & 1 &  1  &  1  &   1   &   1  &  0 \\
\hline
                    \end{tabular}
                \end{normalsize}
            \end{center}
        \end{table*}

In Table \ref{table_course}, Peter passed all compulsory courses, but failed all two elective courses.
John passed one  elective course, but failed  one  compulsory course.
Grace and Jenny passed all compulsory courses and one  elective course,   which met the requirements of the school.
It could be denoted as \{ Grace, Jenny\}$=c_1\wedge c_2\wedge c_3\wedge c_4\wedge (  ec_1\vee ec_2)  $.

\

  Next, we will describe the indefinable granules.

\

\section{The approaching descriptions of indefinable granules}\label{s002}

Based on the definitions of concept-forming operations, we will present unified approaching description methods of
indefinable granules for each of the following types of concepts: formal concept, object-induced three-way concept, object oriented concept  and common-and-necessary concept.

\subsection{The approaching descriptions of $\wedge$-indefinable granules}

Suppose that the target granule $X$
is   $\wedge$-indefinable,
if we can find  $\wedge$-definable granules $X_l$ and $X_u$, such that $X_l\subset X\subset  X_u$, then we say that $X_l$ and $X_u$ are  the approaching descriptions of $\wedge$-indefinable granule $X$ \cite{Li20}.

By Lemma 1.(3), $gf(gf(X))=gf(X)$, thus $gf(X)$ is $\wedge$-definable from Theorem 1.

By Lemma 1.(2), $X\subseteq gf(X)$. Furthermore,
if granule $Y$ is $\wedge$-definable and $X\subset Y$, then  $gf(X)\subseteq gf(Y)=Y$. Thus we get the following.

\textbf{Theorem 5}.  Given a formal context  $(U, A, I)$, $\wedge$-indefinable $X\subset U$   and $f(X) \neq \emptyset$.
  Then $gf(X)$ must be the smallest
$\wedge$-definable granule containing X.

  It is easy to show that here  Theorem 5 is equivalent to  Theorem 4 in \cite{Li20}.

\

Now consider the $\wedge$-definable granules $X_l$, such that $X_l\subset X$.

From a   $(U, A, I)$, we can create a context $(U,  B, J)$, where $a\in A \Leftrightarrow b\in B$, and $(x,a)\in I \Leftrightarrow (x,b)\notin J$. Then \[ X_l=a_1\wedge  a_2  \wedge \cdots  \wedge a_k \Longleftrightarrow U\backslash X_l= b_1 \vee b_2  \vee \cdots  \vee b_k\]

From $X_l\subset X$, we have

\[ b_1 \vee b_2  \vee \cdots  \vee b_k  \supset  U\backslash X   \]

So, to find the largest granule $X_l\subset X$\[ \Longleftrightarrow \min\limits_{b_1 \vee b_2  \vee \cdots  \vee b_k\supset U\backslash X \mbox{\ and\ } (U\backslash X)\bigcap b_i\neq \emptyset, 1\leq i\leq k} \{  b_1, b_2, \cdots  , b_k\in B   \}\]

 \begin{table*}[ht]
            \begin{center}
                \begin{normalsize}
                    \caption{  $(U, A, I)$ in \cite{Li20}}
                    \label{table_tab006}
                    \begin{tabular}
                        {|c|c c  c  c  c|}
                        \hline
                        $U$     & $\ \ \ \ a_1\ \ \ \ \ $       &  $\ \ \ \ a_2\ \ \ \ \ $  &  $\ \ \ \ a_3\ \ \ \ \ $  & $\ \ \ \ a_4\ \ \ \ \ $  &  $\ \ a_5$\ \ \  \\
                       \hline
 1 & 1   & 0  &  0  & 0 &  0  \\

 2 & 1  &  1  & 0 &  1  &  0   \\

 3 & 1   & 0 &  1  &  1  &  1   \\

4 & 1 &  1  &  1  &  0  &   1  \\

5 & 1 &  1  &  1  &  0 &   0  \\

6 & 0 &  1  &  0  &  0 &   1  \\
\hline
                    \end{tabular}
                \end{normalsize}
            \end{center}
        \end{table*}

 \begin{table*}[ht]
            \begin{center}
                \begin{normalsize}
                    \caption{Formal context $(U, B, J)$ from $(U, A, I)$ in \cite{Li20}}
                    \label{table_tab007}
                    \begin{tabular}
                        {|c|c c  c  c  c|}
                        \hline
                        $U$     & $\ \ \ \ b_1\ \ \ \ \ $       &  $\ \ \ \ b_2\ \ \ \ \ $  &  $\ \ \ \ b_3\ \ \ \ \ $  & $\ \ \ \ b_4\ \ \ \ \ $  &  $\ \ b_5$\ \ \  \\
                       \hline
 1 & 0   & 1  &  1  & 1 &  1  \\

 2 & 0  &  0  & 1 &  0  &  1   \\

 3 & 0   & 1  &  0  &  0  &  0   \\

4 & 0 &  0   &  0  &  1  &   0  \\

5 & 0 &  0   &  0  &  1 &   1  \\

6 & 1 &  0   &  1  &  1 &   0  \\
\hline
                    \end{tabular}
                \end{normalsize}
            \end{center}
        \end{table*}

%\newpage

 For example, in  Table \ref{table_tab006}, let us take $X=\{4,5,6\}$, thus we consider  $U\backslash X=\{1,2,3\}$ in  Table \ref{table_tab007}.

 It is easy to see that $\{1,2,3\}= b_2 \vee b_3$  or $\{1,2,3\}= b_2 \vee b_5$.  Therefore,  $X_l=a_2 \wedge a_3=\{ 4,5\}$ or $X_l=a_2 \wedge a_5=\{ 4,6\}$.

\

%For   ($\wedge,  \neg$)-indefinable granules, we can obtain the similar approximation results.

 \
 \subsection{The approaching descriptions of ($\wedge, \wedge, \neg$)-indefinable granules}

Similar to Subsection 3.2,  from a   $(U, A, I)$, we establish a   $(U, A, I, B, J)$, where $a\in A \Leftrightarrow b\in B$, $(x,a)\in I \Leftrightarrow (x,b)\notin J$.

Similar to Subsection 4.1,   we can get the following result.

\textbf{Theorem 6}.  Given a compound context $(U, A, I, B, J)$,  ($\wedge, \wedge, \neg $)-indefinable $X\subset U$,
$f(X) \neq \emptyset$.
  Then $gf(X)$ must be the smallest
($\wedge,  \wedge, \neg$)-definable granule  and $gf(X)\supset X$, where $f$ and $g$ are defined in Subsection 3.2.

  It is easy to show that here  Theorem 6 is equivalent to  Theorem 6 in \cite{Li20}.

Now consider the ($\wedge,  \wedge, \neg$)-definable granules $X_l$, such that $X_l\subset X$. From a compound context $(U, A, I,  B, J)$, we construct a context $(U,  C, K)$, where $a\in A $ or $a\in B\Leftrightarrow c\in C$, and $(x,a)\in I$ or $J \Leftrightarrow (x,c)\notin K$. Then \[ X_l=a_1\wedge  a_2  \wedge \cdots  \wedge a_k \Longleftrightarrow U\backslash X_l= c_1 \vee c_2  \vee \cdots  \vee c_k\]

From $X_l\subset X$, we have

\[ c_1 \vee c_2  \vee \cdots  \vee c_k  \supset  U\backslash X   \]

So, to find the largest granule $X_l\subset X$\[ \Longleftrightarrow \min\limits_{c_1 \vee c_2  \vee \cdots  \vee c_k\supset U\backslash X \mbox{\ and\ } (U\backslash X)\bigcap c_i\neq \emptyset, 1\leq i\leq k} \{  c_1, c_2, \cdots  , c_k\in C   \}\]

 \

 \subsection{The approaching descriptions of $\vee$-indefinable granules}
  Similar to Subsection 3.3,  from a   $(U, A, I)$, we build a   context $(U,   B, J)$, where $a\in A \Leftrightarrow b\in B$, $(x,a)\in I \Leftrightarrow (x,b)\notin J$. Then \[ X=a_1 \vee a_2  \vee \cdots  \vee a_k \Longleftrightarrow U\backslash X= b_1 \wedge b_2  \wedge \cdots  \wedge b_k\]

Similar to Subsection 4.1,  the following new result can be obtained.

\textbf{Theorem 7}.  Given  a  formal context $(U,   B, J)$, $\vee$-indefinable $X\subset U$   and $f(X) \neq \emptyset$.
  Then $U\backslash gf(U\backslash X)$ must be the largest
$\vee$-definable granule contained in $X$.

\

Now consider the $\vee$-definable granules $X_u$, such that $X_u\supset X$.
  Then \[ X_u=a_1\vee  a_2  \vee \cdots  \vee a_k \Longrightarrow
  a_1 \vee a_2  \vee \cdots  \vee a_k  \supset    X   \]

So, to find the smallest granule $X_u\supset X$\[ \Longleftrightarrow \min\limits_{a_1 \vee a_2  \vee \cdots  \vee a_k\supseteq  X \mbox{\ and\ }  X\bigcap a_i\neq \emptyset, 1\leq i\leq k} \{  a_1, a_2, \cdots  , a_k\in A   \}\]

\

\subsection{The approaching descriptions of ($\wedge,\wedge, \vee$)-indefinable granules}

Suppose that the granule $X$ is ($\wedge,\wedge, \vee$)-indefinable over  $(U, A, I, B, J)$ and $f_{CN}(X)  \neq \emptyset$, where $A$ is the set of frequently used attributes, and $B$ is the set of infrequently used attributes.

By definitions of $f_{CN}$ and $g_{CN}$ in Subsection 3.5,  we would obtain $f_{CN}(X)=C\bigcup D$, where $C=\{ a_1, a_2, \cdots  , a_k\} \subseteq A$ and $D=\{b_1, b_2, \cdots  , b_j\} \subseteq B$.

So,
\[g_{CN}(C\bigcup D)= a_1 \wedge a_2  \wedge \cdots  \wedge a_k \wedge  (b_1 \vee b_2  \vee \cdots  \vee b_j)
\]
is ($\wedge, \wedge, \vee$)-definable. Furthermore, if granule $Y$ is ($\wedge, \wedge, \vee$)-definable and $X\subset Y$. Suppose that $f_{CN}(Y)=C'\bigcup D'$, by definitions of $f_{CN}$ and $g_{CN}$ in Subsection 3.5,
$C'\subseteq C$ and $D'\supseteq D$. Thus $Y=g_{CN}f_{CN}(Y)=g(C')\bigcap D'^\diamond\supseteq g(C)\bigcap D^\diamond=g_{CN}f_{CN}(X)$, where $g$ is defined in Definition 2.

  Thus we get the following result.

\textbf{Theorem 8}.  Given a compound context $(U, A, I, B, J)$, ($\wedge, \wedge, \vee$)-indefinable  $X\subset U$ and $f_{CN}(X) \neq \emptyset$.
  Then $g_{CN}f_{CN}(X)$ must be the smallest
($\wedge, \wedge, \vee$)-definable granule   and $g_{CN}f_{CN}(X)\supseteq X$.

\

\

\

\section{Conclusions and future work\label{s5}}

{%\tiny%\scriptsize%\footnotesize

 \begin{table*}[ht]
            \begin{center}
                \begin{footnotesize}
                    \caption{ Granule description definition comparison}
                    \label{table_comparison}
                    \begin{tabular}
                        {|c c  c  |}
                        \hline
                         Granule description type   & Logic formula       &  Condition \ \ \  \\
                       \hline
 $\wedge-definable $ & $X=a_1 \wedge a_2  \wedge \cdots  \wedge a_k$   & $a_1 , a_2  , \cdots  , a_k \in A$   \\

  $(\wedge,\wedge, \neg)-definable $  & $X=a_1 \wedge a_2  \wedge \cdots  \wedge a_k$    & $a_1 , a_2  , \cdots  , a_k \in A\cup B$   \\

  $\vee-definable $  & $X=a_1 \vee a_2  \vee \cdots  \vee a_k$   & $a_1 , a_2  , \cdots  , a_k \in A$  \\

  $(\wedge,\wedge, \vee)-definable $  & $ X=a_1 \wedge a_2  \wedge \cdots  \wedge a_k \wedge  (b_1 \vee b_2  \vee \cdots  \vee b_j)$  &  $a_1 , a_2  , \cdots  , a_k \in A$   \\

  \ \  & \ \   &   and $b_1 \vee b_2  \vee \cdots  \vee b_j\in B$  \\
\hline
                    \end{tabular}
                \end{footnotesize}
            \end{center}
        \end{table*}
}

From Table \ref{table_comparison}, if $X$ is $\vee$-definable, then  $U\backslash X$ is $\wedge$-definable. $(\wedge, \wedge, \neg)$-definable is almost the same as  $\wedge$-definable, yet they have different attribute field.
%The only difference of  $(\wedge, \wedge, \neg)$-definable and $(\wedge,  \neg)$-definable is that they have different condition.
$(\wedge, \wedge, \vee)$-definable is a kind of combination of $\wedge$-definable and $\vee$-definable.

It is common that  the definitions of concept operators are given  in a description language. However, sometimes the logical formula can reflect the essence of the problem.  Through their logical formulas, it is easy to decipher the equivalent relationship between formal concept and object oriented concept, and there
is also a kind of equivalent relationship between formal concept and object-induced three-way concept. Further more, the obvious goal of granule representation is to obtain  an exact logical formula for any definable granule and approaching descriptions for an indefinable granule. In this paper, the logical formulas have been given or derived for all four kinds of concepts. Thus, utilizing the logical relationship among various concepts,  we have derived much concise  and unified equivalent conditions of definable granules and approaching description methods of indefinable granules for all four kinds of concepts. Since it is an NP-hard problem \cite{Li20} to construct the entire concept lattice, our discussions are only based on the definitions of concept-forming operations.

The concept with common attributes has been frequently studied. To investigate the   granules with some special needs, we have proposed one  new  type  of compound concept,
i.e.,  common-and-necessary concept. Its description  is investigated in this paper.

In the future, we will continue to  investigate other kinds of definable granules and explore their exact descriptions or  approaching descriptions for  indefinable granules.  It is also worth to  study the   application of the new type of compound concept in approximate reasoning and cognitive computing.

\

%\noindent\textbf{Acknowledgments} \\
%
%\noindent The authors are grateful to  the referee for useful
%suggestions.

\end{document}